\documentclass[journal]{IEEEtran}
\usepackage{}

\usepackage{multirow}
\usepackage{booktabs}
\usepackage{amsmath}
\usepackage{graphicx}
\usepackage{amssymb}
\usepackage{amsmath,amsthm} % Set the style of Theorem

\usepackage{float}
\usepackage{enumerate}
\usepackage{epstopdf}
\usepackage{dblfloatfix}
\usepackage{caption}
\usepackage{flushend} %%%%% order the bodyText in the last page.
\usepackage{amsfonts}
\usepackage{fancyhdr}
\usepackage{bbm}
\usepackage{cases}
\usepackage{cite}
\usepackage{breqn}
\usepackage{multirow}
\usepackage{mathrsfs}
\usepackage{algpascal}
\usepackage{algc}
\usepackage{algorithmicx}
\usepackage[ruled]{algorithm}
\usepackage{algpseudocode}
\usepackage{graphics}
\usepackage{epsfig}
\usepackage{mathrsfs} % the form of set
\usepackage{dsfont}
\usepackage{color}
\usepackage{setspace}
\usepackage{hyperref}
\usepackage{xcolor}
\usepackage{xpatch}

\floatname{algorithm}{Algorithm}

\algdef{SE}[DOWHILE]{Do}{doWhile}{\algorithmicdo}[1]{\algorithmicwhile\ #1}%

\makeatletter
\def\changeBibColor#1{
\in@{#1}{}
\ifin@\color{red}\else\normalcolor\fi
}

\ifCLASSINFOpdf
 \else
\fi

\begin{document}
\theoremstyle{definition} % Set the style of Theorem
\newtheorem{theorem}{Theorem}[section]
\newtheorem{definition}[theorem]{Definition}
\newtheorem{lemma}[theorem]{Lemma}
\newtheorem{example}[theorem]{Example}
\newtheorem{Proposition}[theorem]{Proposition}
\newtheorem{Corollary}[theorem]{Corollary}

% \title{
% Generative Model-Aided  Continual Learning for CSI Feedback in FDD mMIMO-OFDM Systems 
% }

% \title{\LARGE{
% Generative Model-Aided  Continual Learning for CSI Feedback in FDD mMIMO-OFDM Systems 
% }}
% \title{
% Generative Model-Aided  Continual Learning for CSI Feedback in FDD mMIMO-OFDM Systems 
% }
\title{
Generative Model-Aided  Continual Learning for CSI Feedback in FDD mMIMO-OFDM Systems 
}

\author{
Guijun Liu\textsuperscript{},~\IEEEmembership{}Yuwen Cao\textsuperscript{},~\IEEEmembership{}Tomoaki Ohtsuki\textsuperscript{},~\IEEEmembership{Senior Member, IEEE}, Jiguang He\textsuperscript{},~\IEEEmembership{Senior Member, IEEE}, and Shahid Mumtaz\textsuperscript{},~\IEEEmembership{Senior Member, IEEE} 

\thanks{
This work was supported in part by JST ASPIRE Grant Number JPMJAP2326, Japan. 
G. Liu and Y. Cao are with the
College of Information Science and Technology, Donghua University, Shanghai, China (e-mail: 2232175@mail.dhu.edu.cn, ywcao@dhu.edu.cn). T. Ohtsuki is with the Department of Information and Computer Science, Keio University, Yokohama, Japan (e-mail:ohtsuki@ics.keio.ac.jp). J. He is with  the School of Computing and Information Technology, Great Bay University, Dongguan 523000, China, and Great Bay Institute for Advanced Study (GBIAS), Dongguan 523000, China (e-mail:jiguang.he@gbu.edu.cn).
S. Mumtaz is with the Department of Engineering,
Nottingham Trent University, UK (e-mail:dr.shahid.mumtaz@ieee.org). 
}
}

\maketitle

\begin{abstract}
Deep autoencoder (DAE) frameworks have demonstrated their effectiveness in reducing channel state information (CSI) feedback overhead in massive multiple-input multiple-output (mMIMO) orthogonal frequency division multiplexing (OFDM) systems. However, existing CSI feedback models struggle to adapt to dynamic environments caused by user mobility, requiring retraining when encountering new CSI distributions. Moreover, returning to previously encountered environments often leads to performance degradation due to catastrophic forgetting.  
Continual learning involves enabling models to incorporate new information while maintaining performance on previously learned tasks.
To address these challenges, we propose a generative adversarial network (GAN)-based learning approach for CSI feedback. By using a GAN generator as a memory unit, our method preserves knowledge from past environments and ensures consistently high performance across diverse scenarios without forgetting. Simulation results show that the proposed approach enhances the generalization capability of the DAE framework while maintaining low memory overhead. Furthermore, it can be seamlessly integrated with other advanced CSI feedback models, highlighting its robustness and adaptability.
\end{abstract}

 \vspace{-0.2cm} 

\section{Introduction}

Massive multiple-input multiple-output (mMIMO) has proven to be a crucial technology for the 6th generation (6G) mobile communication system, 
which can enable significant enhancements in system capacity, spectrum efficiency, and data throughput rates\cite{ref2}. However, these benefits of mMIMO systems can be
realized only when the transmitter, especially the base station (BS), has observed accurate downlink channel state information (CSI) \cite{9569752}.
% For time division duplex (TDD) systems, the downlink channel can be obtained from the uplink channel through the channel reciprocity. However, for frequency division duplex (FDD) systems, there is no such channel reciprocity for the downlink channel.
% Thus, the challenge for FDD  mMIMO systems lies in how users efficiently compress and feed back large CSI matrices and how the BS accurately reconstructs them \cite{9569752}.

In frequency division duplex (FDD) mMIMO orthogonal frequency
division multiplexing (OFDM) systems, each antenna requires CSI feedback for all subcarriers, thus leading to significant CSI feedback overhead, as well as excessive consumption of uplink channel bandwidth, power resource, and feedback latency usage. To address the above challenges, CSI feedback compression techniques, such as compressed sensing (CS) \cite{ref5} and codebook-based schemes \cite{ref6}, have been proposed and used in FDD mMIMO-OFDM systems. However, CS-based methods rely heavily on the sparsity assumption of CSI, while codebook-based approaches suffer from increasing computational complexity as the number of antennas grows. When taking the hardware realizations into account, 
% i.e., power consumption, processing capabilities of user equipment (UE's) hardware,
performing accurate CSI estimation and feedback in mMIMO-OFDM is infeasible, especially in low-cost devices. 
% As a result, these methods may become impractical for future 6G mMIMO systems.

Deep neural network (DNN) has been applied in various fields of wireless communications due to its excellent fitting ability. 
% For CSI feedback compression,  Wen \textit{et al.} in \cite{ref7} proposed a novel autoencoder-based framework (CsiNet), which outperforms traditional CS-based methods. In \cite{9296555},
% the authors developed a 
% deep-learning-based CSI compression scheme (DeepCMC) architecture for CSI feedback training. When considering the variation of CSI scenario, 
% reference\cite{ref9} proposed the model transmission method where the encoder and decoder were updated at the user, with the decoder subsequently transmitted to the BS, allowing the deep autoencoder (DAE) model to adapt effectively.
For CSI feedback training, references \cite{ref7}, \cite{9296555},  \cite{9442206}, and \cite{ref9} 
developed novel deep-learning-based CSI compression schemes and model transmission methods.
% where the encoder and decoder were updated at the user, with the decoder subsequently transmitted to the BS. allowing the deep autoencoder (DAE)
% model to adapt effectively.
However, these works ignore the issue of \textit{catastrophic forgetting}, which occurs when the model performs poorly in previously encountered scenarios after adapting to a new environment. This problem leads to repeated data collection and retraining, which significantly degrades communication reliability and system capacity\cite{zxd}.
% Continual learning aims to enable models to retain knowledge across all encountered scenarios \cite{shr}, eliminating the need for retraining from scratch and addressing catastrophic forgetting.
Memory-based learning methods, such as those proposed in \cite{shr,xwc}, rely on data storage for resource allocation and beamforming. However, in mMIMO-OFDM, directly storing large-scale CSI matrices incurs significant storage overhead.
% Meanwhile, various network architectures based on the CsiNet framework have been developed, such as CRNet \cite{ref8}, which employs a multi-resolution architecture to improve the CSI feedback accuracy. 
% Moreover, effectively integrating continual learning with these architectures remains an open challenge.
% Therefore, we claim that for CSI feedback, it is crucial to develop a new CL approach that enhances model generalization while ensuring compatibility with existing methods at minimal cost.

% To address the challenges mentioned above, we draw inspiration from the human learning process, where knowledge is stored in neurons rather than raw data. 
Motivated by the aforementioned challenges, we propose a novel generative adversarial network (GAN)-aided continual learning approach that is tailored for the FDD mMIMO-OFDM for CSI feedback. Specifically, for each scenario, we preserve a GAN generator as a memory unit to capture the CSI distribution of the current environment. In new scenarios, high-quality synthetic data generated by stored generators is leveraged to assist model training.
Notably, in the proposed framework, we design a specialized memory unit to generate high-quality dataset, thus reducing the CSI feedback overhead and making it applicable for \textit{resource-limited} devices.
The main contributions of this letter are summarized as follows:
\begin{itemize}
\item 
% \textcolor{red}{Owing to UE’s mobility, existing DAE models need to be retrained in dynamic wireless environments, thus leading to severe performance degradation due to catastrophic forgetting. However, the proposed method effectively mitigates catastrophic forgetting in CSI feedback, while maintaining low memory overhead and enhancing the generalization capability of the DAE framework.} 
We propose a GAN-aided continual learning approach for the FDD mMIMO-OFDM system to mitigate catastrophic forgetting in CSI feedback, while maintaining  \textit{low memory overhead} and enhancing the generalization capability of deep autoencoder (DAE) frameworks. This is because,  existing DAE models need to be retrained in dynamic wireless environments due to user equipment's (UE’s) movement, leading to severe performance degradation. 
% \item
% It can be seamlessly integrated with advanced DAE models at a low cost, offering inherent robustness.
\item 
To verify its effectiveness, we evaluate the performance of our approach adopting a realistic \textit{3GPP channel model}. In addition, distinct user distribution scenarios are considered in our simulation to verify the effectiveness of our approach. Under each scenario, different CSI distributions are taken into account to further validate the performance of the proposed method. 
\item Simulation results show that the proposed method requires small
storage while achieving excellent CSI feedback performance across encountered scenarios.
% \footnote{{\it Notations}: 
% % Bold uppercase and lowercase letters are used to
% % represent matrices and vectors, respectively. 
% % In addition,
% $\left( {\cdot} \right)^{H}$ and $\left\| {\cdot} \right\|_{\rm{2}}$ denote the complex conjugate and $l_{2}$-norm operations of a matrix, respectively. $\mathbb{C}$ denotes the complex space. $\mathbb{E}\{\cdot\}$ means the expectation of the argument. }
\end{itemize}

% \vspace{-0.2cm}

\section{System Model}
\subsection{FDD mMIMO-OFDM System}
We consider a multi-user mMIMO system in FDD mode, where the BS is equipped with $N_{t} \gg 1$ uniform linear array (ULA) transmit antennas, serving single antenna UEs. OFDM is used over $N_{c}$ subcarriers
for downlink transmission. Over the $n$-th subcarrier, let $x_{n}$ denote the transmitted data symbol, so the received signal $y_{n}$
can be represented as:
\begin{equation} \small\label{equ:1}
y_{n}=\mathbf{h}_{n}^H\mathbf{v}_{n}x_{n}+z_{n},
\end{equation}
where $\mathbf{h}_n\in\mathbb{C}^{N_t\times1}$ corresponds to the channel vector, $\mathbf{v}_{n}\in\mathbb{C}^{N_t\times1}$ denotes the precoding vector and $z_{n}$ is the additive noise. The entire channel matrix estimated by UE can be denoted as
$
\mathbf{H} = [\mathbf{h}_1, \mathbf{h}_2, \ldots, \mathbf{h}_{N_c}] \in \mathbb{C}^{N_t \times N_c}.
$
In mMIMO systems, the channel matrix is of high dimension, which should be compressed and then fed back to the BS to avoid large CSI feedback overhead, as well as the excessive uplink channel bandwidth consumption.
 \vspace{-0.4cm}

\subsection{Deep Autoencoder CSI Feedback Framework}

\begin{figure}[t!] 
    \centering
    \setlength{\abovecaptionskip}{-0.1cm}
   \setlength{\belowcaptionskip}{-0.6cm} 
    \includegraphics[width=7cm]{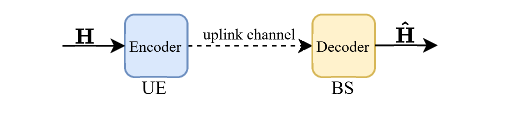}
    \caption{An overview of the autoencoder CSI feedback framework.
    } 
    \label{fig:fignew1}
\end{figure}

As shown in Fig. \ref{fig:fignew1}, 
to reduce the CSI feedback overhead, DAE is adopted to compress CSI and then recover it. 
 The mMIMO CSI matrix $\mathbf{H}$ is firstly compressed into low-dimensional codewords $\mathbf{s}$ via the encoder network. Afterward, the codewords are fed back to the BS as shown in (\ref{equ:3}):
\begin{equation}
\small
\label{equ:3}
\mathbf{s}=f_{enc}(\mathbf{H}),
\end{equation}
where $\mathbf{s} \in \mathbb{C}^{V \times 1}$ denotes the compressed codewords, $V$ represents the size of the codewords, 
and $f_{enc}(\cdot)$ refers to the encoder neural network. 
The compressed ratio can then be denoted as $\gamma =\frac{V}{2N_{t}N_{c}}$.

The BS utilizes the decoder network to recover the original channel matrix $\mathbf{H}$ based on the following decoding criterion:
\begin{equation}
\small
\label{equ:4}
\hat{\mathbf{H}}=f_{dec}(\mathbf{s}),
\end{equation}
where $
\hat{\mathbf{H}} \in \mathbb{C}^{N_t \times N_c}
$ denotes the recovered CSI matrix and 
$f_{dec}(\cdot)$ denotes the decoder neural network. The used mean square error (MSE) loss function is given by
\begin{equation}
\small
\label{equ:6}
Loss=\frac{1}{N}\sum\nolimits_{i=1}^{N}||\mathbf{H}_{i}-\hat{\mathbf{H}}_{i}||_{2}^{2}, 
\end{equation}
where \( N \) denotes the total number of data samples, 
\( \mathbf{H}_i \) and \( \hat{\mathbf{H}}_i \) represent the \( i \)th data and recovered CSI data, respectively. Herein, $\left\| {\cdot} \right\|_{\rm{2}}$ means the $l_{2}$-norm operation of a matrix. 

\begin{figure}[t]
% \vspace{-0.2cm} % 向上移动 0.5cm
     \setlength{\belowcaptionskip}{-0.6cm}    \setlength{\abovecaptionskip}{-0.1cm}
    \centering
 \includegraphics[width=9.2cm]{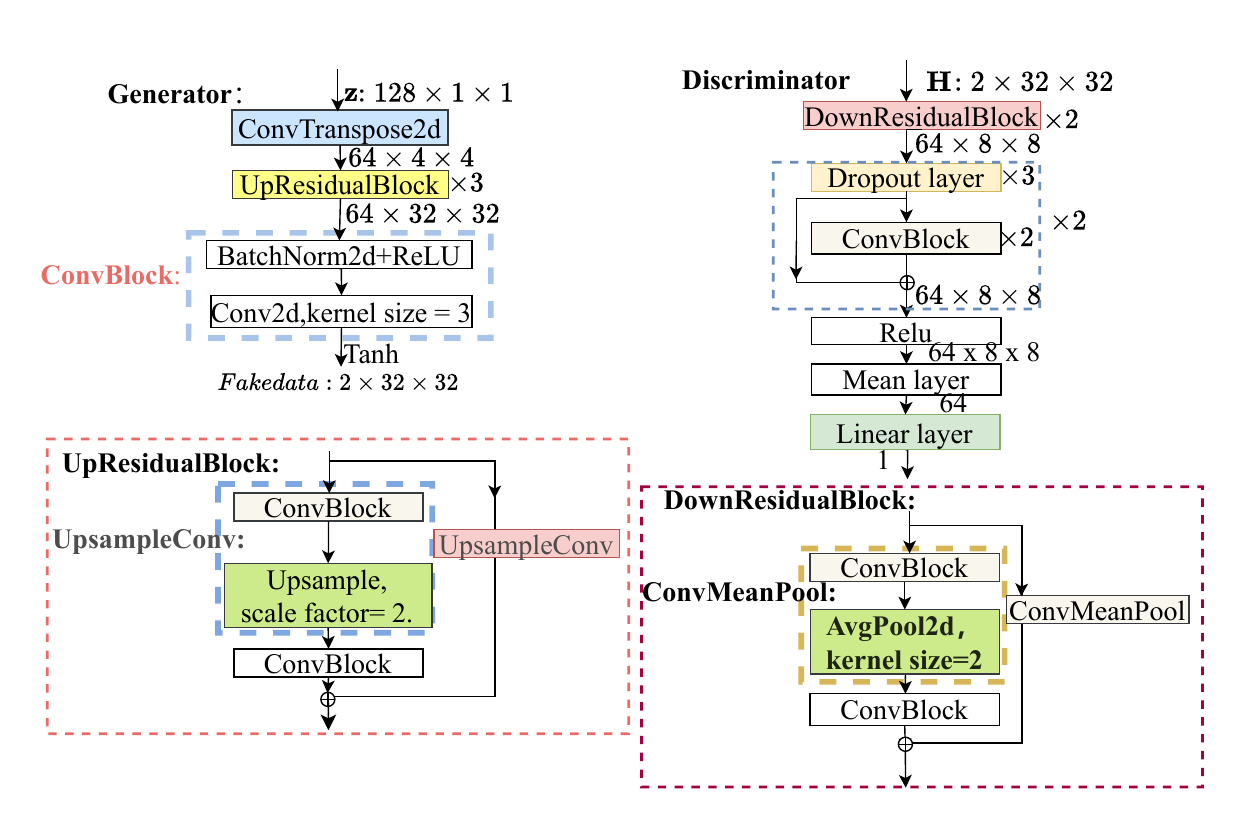}
    \caption{The adopted GAN model structure, which consists of a generator $G(\cdot)$ and a discriminator $D(\cdot)$. 
    The ``Upsample, factor=2" module refers to the nearest neighbor interpolation with a scale factor of 2, while ``Mean Layer" denotes an averaging operation applied to the last two dimensions of the input.
    }
    \label{fig:fig1}
\end{figure} 

\textit{Remark 1}: The DAE's strong fitting ability results in a very high precision of the restored $\hat{\mathbf{H}}$ \cite{ref9}.
However, in practical wireless communication scenarios, significant variations may occur due to \textit{UE's mobility}. Such DAE structure lacks generalization. Notice that the continuous online learning approach in \cite{ref9} focuses on storing previous model parameters via bilevel optimization, while our method employs GAN generators as memory units to capture \textit{CSI distributions}. Such design inherently reduces memory overhead, as shown in \textbf{Table \ref{table2}}. In addition, the work \cite{ref15} leverages memory to store key data for facilitating robust beamforming in dynamic channel environment, while our method employs GAN generators as memory units to capture CSI distribution.\vspace{-0.2cm}

% \begin{figure*}
%     % \vspace{-0.3cm} % 根据需要调整此
%          \setlength{\belowcaptionskip}{-0.4cm}    \setlength{\abovecaptionskip}{-0.0cm}
%     \centering
%     % \setlength{\abovecaptionskip}{-0.1cm}
% %\setlength{\belowcaptionskip}{-0.1cm} 
%     \includegraphics[width=10cm]{fig/vtc3.drawio.pdf}
%     \caption{
% The proposed continual learning framework for CSI feedback in dynamic environments. The dashed arrow indicates that in the encountered scenarios, the memory unit $\mathcal{M}$ has already captured the data distribution \( data_{0, \ldots, t} \)
% through GAN-based channel modeling. 
% By employing the memory unit, it is sufficient to collect only the \( data_t \) in the current scenario \( t \) to address the issue of catastrophic forgetting.
% } 
%     \label{fig:fig3}
% \end{figure*} 

% \titlespacing*{\subsection}{0pt}{5pt}{5pt}
\section{The Proposed Continual Learning Framework}
In this section, we first introduce the adopted GAN model.
Then, we present how to apply the GAN model as a special memory unit to solve the catastrophic forgetting problem in CSI feedback scenarios. \vspace{-0.2cm}

\subsection{The Rationale of The Generative Adversarial Network}
In the context of mMIMO-OFDM systems, storing data from each scenario is inefficient due to the large size of CSI matrices.
% By contrast, humans retain only the essential information they understand during learning, rather than raw data. 
Motivated by the challenge above and inspired by the exceptional channel modeling capabilities of GAN proven in \cite{ref15}, we propose using a GAN model to capture the approximate channel distribution of encountered scenarios, which assists continual learning for the CSI feedback model.
% \footnote{
% The GAN is adopted for its efficiency in generating high-fidelity CSI samples with low computational cost, critical for real-time mMIMO-OFDM systems. Compared to \cite{9442206}, our GAN incorporates residual blocks and Earth Mover (EM) distance loss given in (\ref{equ:7}) to better model \textit{spatial correlations} in mMIMO-OFDM channels, enhancing synthetic data quality without increasing complexity. Note that our framework differs from prior research \cite{ref2} by tailoring GAN generators to mMIMO channel characteristics and integrating them directly into the CSI feedback loop, ensuring efficient adaptation to dynamic environments.
% }

As shown in Fig. \ref{fig:fig1}, GAN consists of a generator and a
discriminator. The generator inputs normal distribution data $\mathbf{z}$
and then generates fake CSI data $\mathbf{\tilde{H}}$
based on the channel distribution
$P_{f}(\tilde{\mathbf{H}})$ that it captures. 
The discriminator is used to determine
whether the input CSI is real data $\mathbf{H}$
collected by UEs or fake data $\mathbf{\tilde{H}}$
generated by the generator. 
The $Loss$ for GAN training can be understood as the discriminator's accuracy in determining whether the CSI is true or fake. Using alternating training, the generator tries to generate data similar to true CSI, i.e., maximize the $Loss$, while the discriminator tries to judge the truth of the data, i.e., minimize the $Loss$. This adversarial approach will eventually allow the generator to approximately model the real CSI distribution $P_{r}(\mathbf{H})$.

\begin{figure}
    % \vspace{-0.3cm} % 根据需要调整此
         \setlength{\belowcaptionskip}{-0.4cm}    \setlength{\abovecaptionskip}{-0.0cm}
    \centering
    \hspace*{-0.8cm}
    \includegraphics[width=7cm]{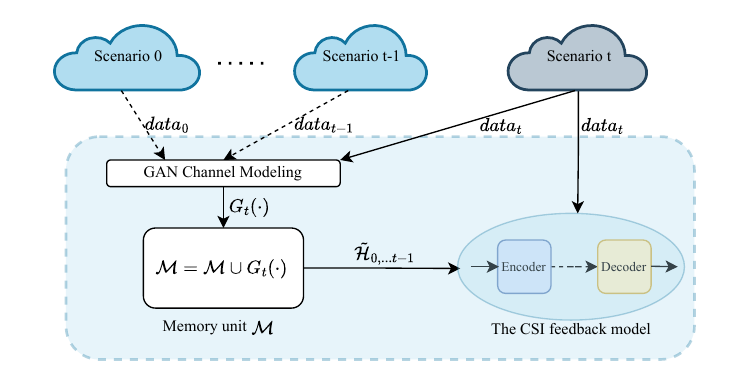}
    \caption{
The proposed continual learning framework for CSI feedback in dynamic environments. The dashed arrow indicates that in the encountered scenarios, the memory unit $\mathcal{M}$ has already captured the data distribution \( data_{0, \ldots, t} \)
through GAN-based channel modeling. 
% By employing the memory unit, it is sufficient to collect only the \( data_t \) in the current scenario \( t \) to address the issue of catastrophic forgetting.
} 
    \label{fig:fig3}
\end{figure} 

To ensure convergence and improve the accuracy of model training, we adopt the EM distance as the loss function. The EM distance $Loss$ can be denoted as:
\begin{equation}
\small
\label{equ:7}
Loss(\mathbf{H},\tilde{\mathbf{H}})=\sup_{f\in\text{1-Lip}}\mathbb{E}_{\mathbf{H}\sim P_r(\mathbf{H})}\{f(\mathbf{H})\}-\mathbb{E}_{\tilde{\mathbf{H}}\sim P_{f}(\tilde{\mathbf{H}})}\{f(\tilde{\mathbf{H}})\}, 
\end{equation}
where $\text{1-Lip}$ denotes the 1-Lispschitz function and $f(\cdot)$ denotes the discriminator network. 
Compared to the direct classification loss, the EM distance adds the 1-Lipschitz function constraint to the model, making the model training process more convergent when the difference between the distributions of $P_r(\mathbf{H})$ and $P_{f}(\tilde{\mathbf{H}})$  is significant
and therefore enhancing modeling performance. 

Let $G(\cdot)$ and $D(\cdot)$ be the generator and the discriminator, respectively. Eq. (\ref{equ:7}) can be implemented as Eq. (\ref{equ:ge}) for the generator
\begin{equation}
\small
\label{equ:ge}
Loss(\boldsymbol{z})=-D(G(\boldsymbol{z})),
\end{equation}
and Eq. (\ref{equ:8}) for the discriminator using consistency regularization GAN (CTGAN) \cite{gan}
\begin{equation} 
\small
\label{equ:8}
\begin{aligned}
Loss(\mathbf{H},\tilde{\mathbf{H}})=&\mathbb{E}_{\boldsymbol{z}\sim N(0,\mathbf{I})}\left\{D(G(\boldsymbol{z}))\right\}-\mathbb{E}_{\mathbf{H}\sim P_r(\mathbf{H})}\{D(\mathbf{H})\}
\\&+\lambda_1\mathbb{E}_{\widehat{\mathbf{H}}}\{\big(\|\nabla_{\widehat{\mathbf{H}}}D(\widehat{\mathbf{H}})\|_2-1\big)^2\}+\\
&\lambda_2\mathbb{E}_{\mathbf{H}\sim P_{r}(\mathbf{H})}\{\max\left(0, \|D_{1}(\mathbf{H})-D_{2}(\mathbf{H}) \|_2 -M^{\prime}\right)
\},
\end{aligned}
\end{equation}
where $\widehat{\mathbf{H}}=i\mathbf{H}+(1-i)G(\mathbf{z})$, $i$ follows uniform distribution.   
$\nabla_{\widehat{\mathbf{H}}}D(\widehat{\mathbf{H}})$ 
denotes the gradient of $\widehat{\mathbf{H}}$, $D_{1}(\mathbf{H})$ and $D_{2}(\mathbf{H})$ denote the outputs of the discriminator when two dropout probabilities are applied to $D(\cdot)$ when $\mathbf{H}$ is input. $D_{1}^{\prime}(\mathbf{H})$ and $D_{2}^{\prime}(\mathbf{H})$ represent the output of the second-to-last layer of $D(\cdot)$ with dropout layers. 
$\lambda_1$, $\lambda_2$, and $M^{\prime}$ are hyperparameters. %\vspace{-0.6cm}
%\vspace{-0.3cm}

% \titlespacing*{\subsection}{0pt}{5pt}{5pt}
\subsection{The Proposed Continual Learning CSI Feedback Method}

For dynamically changing wireless communication scenarios, as shown in Fig. \ref{fig:fig3}, our proposed method addresses new scenarios (scenario \( t \)) by leveraging a memory unit $\mathcal{M}$. $\mathcal{M}$ is formed by using GAN-based channel modeling to capture the CSI data distributions from previously encountered scenarios and storing each \( G_t(\cdot) \) as a specialized memory. Combining with Fig. \ref{fig:fig1}, for mMIMO-OFDM scenarios, storing the convolution-based \( G_t(\cdot) \) requires low overhead compared to storing raw CSI data. The memory unit $\mathcal{M}$ generates synthetic CSI data set $\tilde{\mathcal{H}}_{0,\ldots,t-1}$ that encapsulates information from past scenarios ($0,\ldots,t-1$):
\begin{equation} \label{equ:19}
\tilde{\mathcal{H}}_{0,\ldots, t-1}=G_{0,\ldots, t-1}(\boldsymbol{z}),  \;\;
 \boldsymbol{z}\sim N(0,\mathbf{I}).
\end{equation}
The synthetic $\tilde{\mathcal{H}}_{0,\ldots,t-1}$, combined with the current $data_{t}$,
is then used to train the CSI feedback model as (\ref{equ:20}):
\begin{equation} \label{equ:20}
\hat{\mathbf{H}}=f_{dec}(f_{enc}(\mathbf{H})), \mathbf{H} \in \{ 
\tilde{\mathcal{H}}_{0,\ldots,t-1} \cup data_{t}
\}. 
\end{equation}
The CSI feedback model is trained on a dataset that incorporates all previously encountered CSI information, ensuring excellent performance across encountered wireless communication scenarios and enhancing the model's generalization capability with low  storage cost.
Additionally, \( G_{t}(\cdot) \) derived from modeling \( data_t \) is updated into the memory unit $\mathcal{M}=\mathcal{M} \cup G_{t}(\cdot)$ to retain memory of the current environment.   
% \footnote{\textcolor{red}{As mentioned by reference \cite{cao2025memristor} that it is feasible to mitigate the storage burden by storing special memory unit. 
% To ensure the robustness of the algorithm, it is expected to store the convolution-based $G_t(\cdot)$ for as many scenarios as possible. However, storage burden will be involved for the whole system. How to appropriately store $G_t(\cdot)$ for many scenarios while maintaining memory overhead will be studied in our future research.}}

% Notably, the proposed method is compatible with other state-of-the-art (SOTA) DAE, because it focuses on designing a specialized memory unit to generate high-quality dataset and is decoupled from the autoencoder structure. As a result, the proposed method can be seamlessly integrated with other SOTA CSI feedback DAE structure.

\subsection{Complexity Analysis}
The computational complexity of the proposed GAN-aided continual learning framework for CSI feedback is incurred mainly by the GAN model training operation. Specifically, the computational complexity order of the GAN training is \(\mathcal{O}(N_{t}N_{c})\). 
% Accordingly, the proposed GAN-aided continual learning framework requires an $
% \mathcal{O}\left(N_{t}N_{c} \right)$
% computational complexity.  
During the inference stage, let the number of generated models be denoted by $N_g$. Since the generator $G(\cdot)$ in GANs is predominantly composed of convolutional operations, the computational complexity of the inference process across all models scales as $\mathcal{O}(N_g N_{t} N_{c})$.
% \footnote{\textcolor{red}{
% Note that the knowledge-driven meta-learning method proposed in \cite{10308721} requires an
% $\mathcal{O}(N_{\mathrm{us}}N_{t}\lceil\beta M_{m}\rceil(\lceil\alpha L_{m}\rceil+N_{\mathrm{sb}}))$ complexity cost for meta training, and imposes an  
% $\mathcal{O}(\tilde{N}_{\mathrm{ua}}N_{\mathrm{sb}}N_{t}^2(N_{{\mathrm{gran}}}+N_{t}))$ +$\mathcal{O}(\widetilde{N}_{\mathrm{ud}}N_{\mathrm{t}}^3)$ complexity cost for target retraining. $N_{\mathrm{us}}$, $\widetilde{N}_{\mathrm{ua}}$, and $\widetilde{N}_{\mathrm{ud}}$ are the product of the number of UEs and slots for all CSI tasks, the maximum number of channel samples of all UEs across all CSI scenarios encountered, and the product of the number of UEs and the delay spread, respectively. $N_{\mathrm{sb}}$ and $N_{{\mathrm{gran}}}$ represent the number of sub-bands and the number of subcarriers within each sub-band. $L_{m}$ and $M_{m}$ mean the degree of feature diversity in spatial and frequency domains, respectively. $\alpha$ and $\beta$ are set to scale the diversity of feature of each slot.  
% }
% }
% \textcolor{red}{
% Note that the knowledge-driven meta-learning method proposed in \cite{10308721} incurs computational cost mainly for the meta training and the target retraining.
% }
Note that the memory storage consumption of the proposed framework will be analyzed in the following Section \ref{Sim:nmse}.

\begin{figure}[t!] 
    \centering
   \setlength{\belowcaptionskip}{-0.5cm}  
     \includegraphics[width=5cm,height=2.5cm]{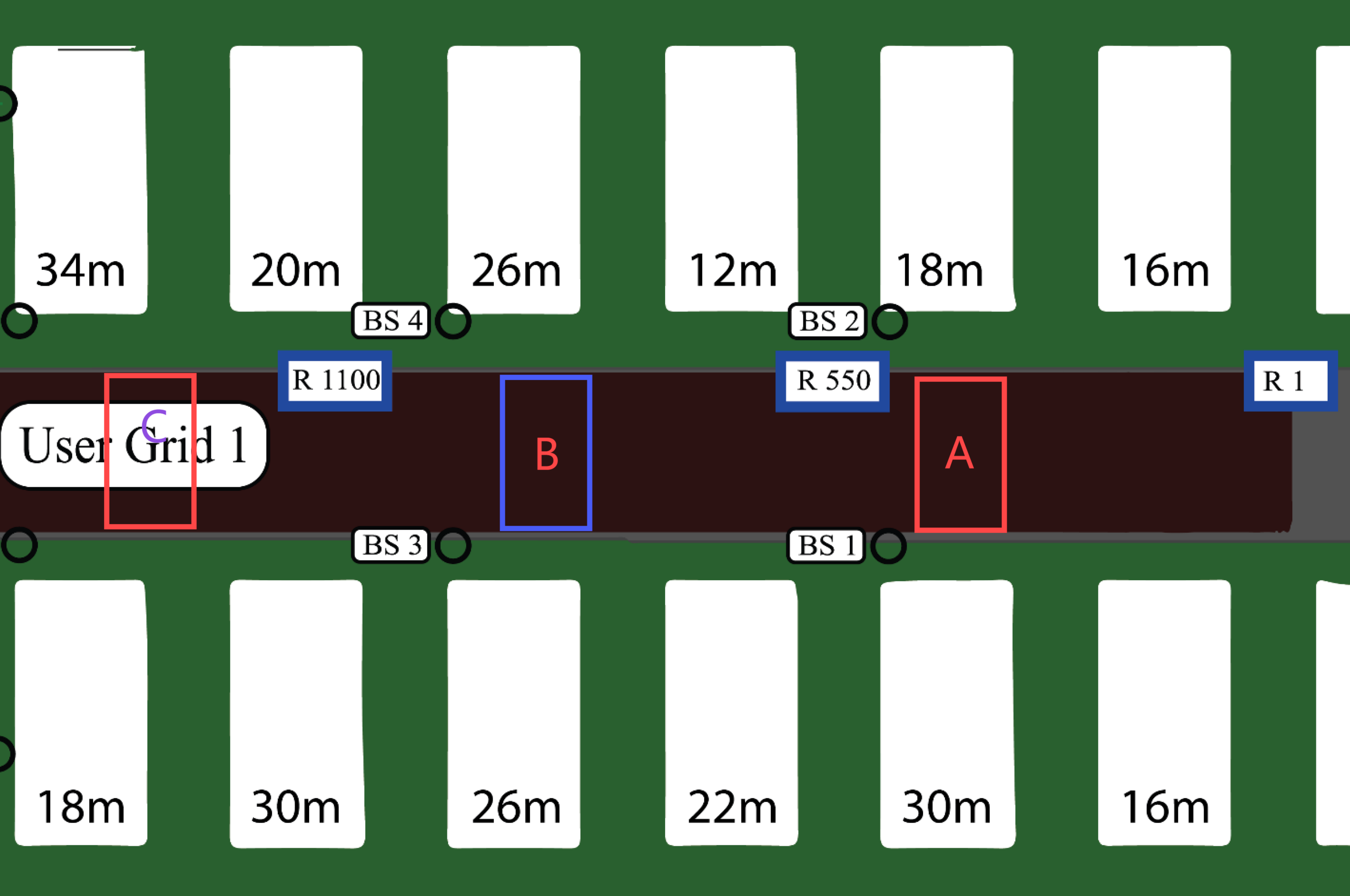}
    \caption{“O1\_28" an outdoor scenario of two streets and one intersection at operating frequencies 28 GHz in \textit{DeepMIMO}.
    } 
    \label{fig:fig4}
\end{figure}

% \titlespacing*{\subsection}{0pt}{5pt}{5pt}

\section{Simulation Results}
\subsection{Simulation Specification}
\label{Simu:spe}
%
% In this section, we introduce hyperparameters and dataset generation settings in our experiments.
In this subsection, we introduce the hyperparameter setting of the GAN training model, as well as dataset generation settings in our experiments. We implement our proposed GAN-aided continual learning framework on the generated \textit{spacial-frequency domain} channel.

\begin{enumerate}
    \item \textit{Hyperparameters Settings}: 
    In the proposed method, we set the learning rate to 0.001, the compression ratio \(\gamma\) to 1/16, \(\lambda_1\) to 10, \(\lambda_2\) to 2, and \(M^\prime\) to 0.2. The dropout probability is 0.5, and the training epoch is 300 with a batch size of 100. Adam is used as the optimizer.
    
    \item \textit{Dataset Settings}:
    \textit{DeepMIMO} is a publicly available dataset generated by the Remcom Wireless Insite tool\cite{deepmimo}. 
    As illustrated in Fig. \ref{fig:fig4}, we design three distinct scenarios to verify the effectiveness of our method:
    \begin{enumerate}
        \item Scenario A: UEs located within rows 500 to 534.
        \item Scenario B: UEs located within rows 950 to 984.
        \item Scenario C: UEs located within rows 1300 to 1334.
    \end{enumerate}
\end{enumerate}
 For all scenarios, BS4 is selected as the corresponding BS. In addition, we set the bandwidth to be 0.05 GHz, the transmitting antennas to be 32, the number of subcarries to be 32, and the number of paths to be 25.
In addition, we utilize the first 6000 CSI data samples from each scenario as the dataset, with 5000 samples allocated for training and 1000 for testing. The data is normalized to the range \([-1, 1]\). Initially, we adopt the original CsiNet \cite{ref7} model as the CSI feedback structure. 
Besides, the normalized MSE (NMSE) is adopted as the metric:
\begin{equation}
\small
NMSE=\mathbb{E}\left\{\frac{\|\widehat{\mathbf{H}}-\mathbf{H}\|_2^2}{\|\mathbf{H}\|_2^2}\right\},\end{equation}
where $\mathbb{E}\{\cdot\}$ represents the expectation of the argument. \vspace{-0.4cm}

\subsection{NMSE Performance Evaluation}
\label{Sim:nmse}

\begin{table}[t!]
\centering
\caption{NMSE (dB) performance of the proposed scheme compared to the MTL scheme across epidemic scenarios.}
\small % 设置表格字体为小号
\scalebox{0.65}{ % 将表格缩小为原来的80%

\begin{tabular}{|c|c|c|c|c|c|}
\hline
\multirow{2}{*}{$K$} & \multirow{2}{*}{Method} & \multirow{2}{*}{\begin{tabular}{c}
After training \\ on
\end{tabular}} & \multicolumn{3}{c|}{NMSE (dB)} \\ \cline{4-6}
 &  &  & A & B & C \\ \hline
\multicolumn{3}{|c|}{MTL} & \textbf{-24.49} & \textbf{-21.23} & \textbf{-26.78} \\ \hline
% \multicolumn{3}{|c|}{\textcolor{red}{ML}} & \textcolor{red}{\textbf{-22.34}} & \textcolor{red}{\textbf{-19.43}} & \textcolor{red}{\textbf{-24.76}} \\ \hline
\multirow{3}{*}{1000} & \multirow{3}{*}{\begin{tabular}{c}
Proposed \\ method
\end{tabular}} & A & -25.34 & - & - \\ \cline{3-6}
 &  & B & -16.28 & -19.76 & - \\ \cline{3-6}
 &  & C & -17.74 & -14.64 & -25.24 \\ \hline
\multirow{3}{*}{2000} & \multirow{3}{*}{\begin{tabular}{c}
Proposed \\ method
\end{tabular}} & A & -25.34 & - & - \\ \cline{3-6}
 &  & B & -18.31 & -19.59 & - \\ \cline{3-6}
 &  & C & -19.43 & -16.11 & -25.52 \\ \hline
\multirow{3}{*}{5000} & \multirow{3}{*}{\begin{tabular}{c}
Proposed \\ method
\end{tabular}} & A & -25.34 & - & - \\ \cline{3-6}
 &  & B & -20.86 & -19.15 & - \\ \cline{3-6}
 &  & C & -20.85 & -17.88 & -25.37 \\ \hline
\multirow{3}{*}{10,000} & \multirow{3}{*}{\begin{tabular}{c}
Proposed \\ method
\end{tabular}} & A & -25.34 & - & - \\ \cline{3-6}
 &  & B & -22.81 & -19.38 & - \\ \cline{3-6}
 &  & C & \textbf{-22.49} & \textbf{-18.61} & \textbf{-24.47} \\ \hline
\end{tabular}
}
\label{tab:nmse_results}\vspace{-0.2cm}
\end{table}

\begin{table}[t!]
\centering
%\small % 设置表格字体为小号
\footnotesize
\caption{The memory storage size in the current continual learning experiment is presented, assuming each element is stored as 4 bytes.}
\scalebox{0.8}{ % 将表格缩小为原来的80%
\resizebox{\linewidth}{!}{
\begin{tabular}{|c|c|c|c|}
\hline \textbf{Method} & Proposed & MinMax/Reservoir & Joint \\
\hline \textbf{Memory cost} & 3.552 M & 15.625 M & 78.125 M \\
\hline
\end{tabular}}}
\label{table2}\vspace{-0.4cm}
\end{table}

\begin{figure}[t!] 
% \vspace{-0.4cm} % 向上移动 0.5cm
    \centering
      \setlength{\belowcaptionskip}{-0.6cm}    \setlength{\abovecaptionskip}{-0.0cm}
    \includegraphics[width=7.5cm]{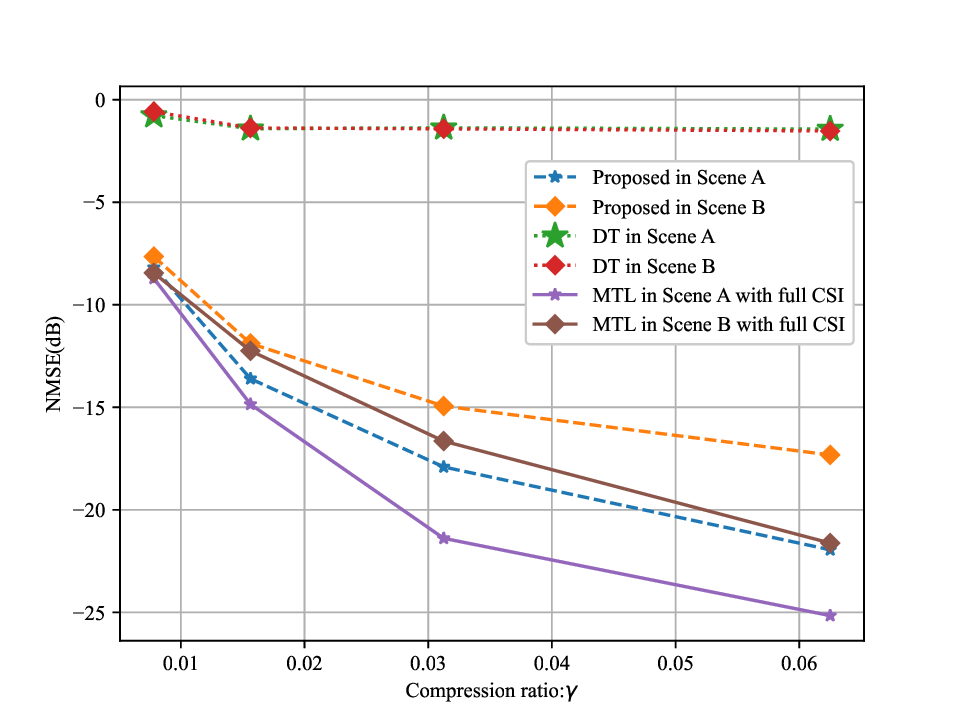}
    \caption{NMSE (dB) performance comparison of the proposed approach, DT, and MTL under various compression ratios $\gamma$.}
    \label{fig:simulation2}
\end{figure}

\begin{table}[t]
\centering
% \small % 设置表格字体为小号
\footnotesize
\caption{Comparison of NMSE (dB) performance across different network structures.}
\scalebox{0.8}{ % 将表格缩小为原来的80%
\resizebox{\linewidth}{!}{
\begin{tabular}{|c|c|c|c|c|}
\hline
\multirow{2}{*}{\textbf{Network structure}} & \multirow{2}{*}{\textbf{After training on}} & \multicolumn{3}{|c|}{\textbf{NMSE (dB)}} \\ 
\cline{3-5} 
& & A & B & C \\ 
\hline
\multirow{3}{*}{CsiNet\cite{ref7}} & A & -25.34 & - & - \\ 
\cline{2-5}
& B & -22.81 & -19.38 & - \\ 
\cline{2-5}
& C & -22.49 & -18.61 & -24.47 \\ 
\hline
\multirow{3}{*}{CRNet\cite{ref8}} & A & -28.74 & - & - \\ 
\cline{2-5}
& B & -26.44 & -22.06 & - \\ 
\cline{2-5}
& C & -25.71 & -19.89 & -27.07 \\ 
\hline
\end{tabular}
}
\label{table3}
}
\end{table}

We configure the model to sequentially learn the three wireless communication scenarios in the order of A, B, and C.
Firstly, we compare the proposed method with the multi-task learning (MTL), as studied in \cite{zxd}.  
MTL refers to a model trained in advance using all the data collected from the three scenarios and is considered as the upper bound for continual learning.
Let \( K \) represent the number of fake data generated by each generator in the memory unit $\mathcal{M}$ when facing a new scenario. Note that the $K$ generated samples are temporary, and solely used during training iterations, and are not permanently stored.\footnote{Our memory cost analysis (as given in \textbf{Table \ref{table2}}) specifically reflects permanent storage of generator parameters, emphasizing that no additional \textit{long-term storage} is needed for synthetic data. This distinguishes our approach from the currently existing methods requiring persistent data retention.} As shown in \textbf{Table \ref{tab:nmse_results}}, when \( K = 10,000 \), the proposed method achieves NMSE performance that is approximately comparable to MTL in all the scenarios.
% After training on three scenarios, 
% we then set the compressed ratios  $\gamma$ to be
% 1/16, 1/32, 1/64, and 1/128, respectively. 
After training on three scenarios, we set the compression ratios $\gamma$ to be 1/16, 1/32, 1/64, and 1/128. We then compare performance in scenarios A and B with the direct transfer (DT) scheme \cite{ref9}, which adapts online without memory, and the MTL scheme. We observe from Fig. \ref{fig:simulation2} that, with given compression ratios $\gamma$, the NMSE gap between our approach and MTL remains
consistently small
across different scenarios. 
% \textcolor{red}{This figure also shows that our method achieves superior NMSE performance than CRNet that adapts with memory for CSI feedback.}
However, in practical wireless environments, it is often impractical to obtain the complete data set required by MTL. Therefore, we deduce that our scheme is more suitable for such dynamic scenarios.%\vspace{-0.3cm}

\begin{figure}[t!] 
\vspace{-0.4cm} % 向上移动 0.5cm
  \setlength{\belowcaptionskip}{-0.6cm}    \setlength{\abovecaptionskip}{-0.0cm}
    \centering 
    \includegraphics[width=8.6cm]{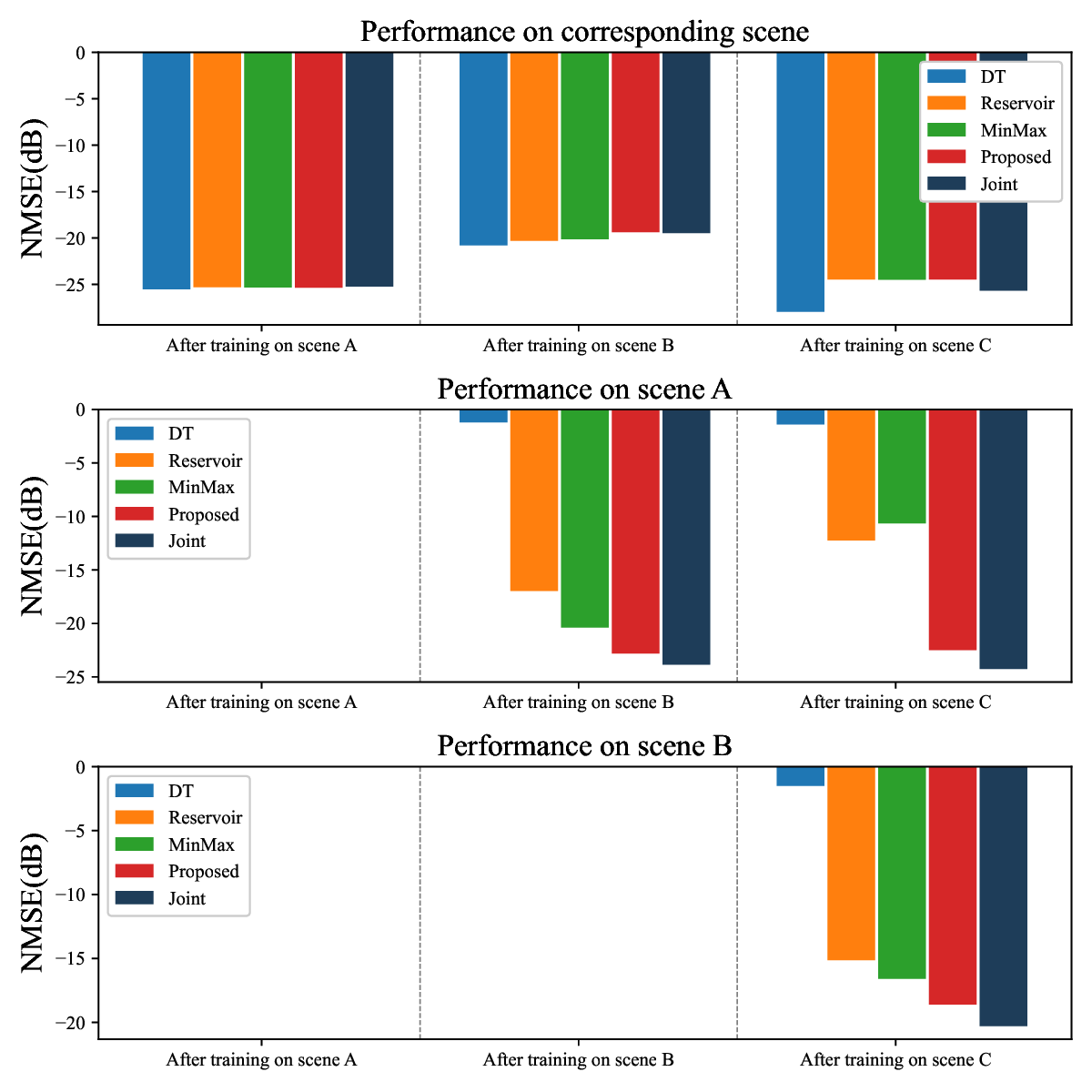}
    \caption{Comparison of NMSE (dB) performance among different memory-based methods.
    } 
    \label{fig:simulation}
\end{figure}

Next, we compare the performance of our method with four other approaches: i) DT \cite{ref9};
%Direct transfer (DT), which operates without a memory unit and performs online adaption in new scene directly, similar to \cite{ref9};
ii) Reservoir sampling (Reservoir)\cite{xwc}, where 2000 samples are uniformly selected at each step to form the memory unit; iii) MinMax \cite{shr}; 
% which prioritizes samples with larger losses during training and memory selection for greater fairness, with the memory unit containing 2000 samples; 
iv) Joint learning (Joint), which accumulates and stores accumulative data \( data_{0,\ldots,t} \) for model training. As shown in Fig. \ref{fig:simulation}, 
direct transfer achieves excellent NMSE performance in new scenarios but suffers from catastrophic forgetting, performing poorly in previously encountered ones. 
% The Reservoir and MinMax methods partially mitigate the forgetting issue; however, selecting and storing data in the context of CSI feedback leads to performance degradation and is not memory-efficient. 
In contrast, the proposed method, which only stores the parameters of GAN generators, closely approximates the performance of the Joint method that stores all accumulated data.
As shown in \textbf{Table \ref{table2}}, the proposed method is more memory-efficient. It is also worth noting that for more larger-scale antenna arrays, the space-saving advantage of the convolution-based GAN generator structure will become even more significant, making it particularly suitable for mMIMO-OFDM CSI feedback systems.

% In the above experiments, the DAE structure we used is the original CsiNet. However, many advanced DAE structures have already been proposed. 
% As described in Section III, the focus of our method is to introduce a novel memory mechanism tailored for CSI feedback scenarios. Our method is not restricted to a specific DAE structure, making it compatible with more advanced DAE designs to further enhance CSI feedback performance.
Furthermore, we integrate our method with CRNet \cite{ref8} to demonstrate its robustness.
% CRNet adopts the cosine annealing learning rate schedule as described in \cite{ref8},
% with an initial learning rate set to be 0.1. 
\textbf{Table \ref{table3}} shows that our method combined with CRNet further improves NMSE performance for CSI feedback. 
%Using channel matrices in their original domain demonstrates our method’s generality across input representations. While angle-delay transformation can exploit sparsity, our results in \textbf{Table \ref{table3}} show robust performance without such preprocessing, validating adaptability to diverse system setups.

\section{Conclusion}
To address the catastrophic forgetting problem of CSI feedback models in dynamic FDD mMIMO-OFDM systems, we proposed a novel GAN-aided continual learning framework to enhance the generalization ability of existing DAE techniques. 
% Furthermore, the proposed method is not confined to any specific CSI feedback model, which allows it to be easily adapted to various models, thereby ensuring its robustness. 
In our framework, we preserve a GAN generator as a memory unit to capture the CSI distribution of the current environment. In new scenarios, high-quality synthetic data generated by stored generators is leveraged to assist model training. Simulation results demonstrated that the proposed method requires low storage while achieving excellent CSI feedback performance across encountered scenarios, effectively mitigating the catastrophic forgetting issue in dynamic environments.

\section{Acknowledgments}
We thank Qwen for assistance with language polishing.
 
\bibliographystyle{IEEEtran}%
\bibliography{liufile}
\end{document}